# XAI-based gait analysis of patients walking with Knee-Ankle-Foot orthosis using video cameras

Arnav Mishra[1], Aditi Shetkar [2], Ganesh M. Bapat[3], Rajdeep Ojha[4], Tanmay Tulsidas Verlekar[2]

[1] Dept. of EEE, BITS Pilani, K K Birla, Goa Campus, 403726 Goa, India

[2] Dept. of CSIS, BITS Pilani, K K Birla, Goa Campus, 403726 Goa, India

[3] Dept. of Mech Eng, BITS Pilani, K K Birla, Goa Campus, 403726 Goa, India

[4] Dept. of Physical Medicine and Rehabilitation, Christian Medical College Vellore, Tamil Nadu, India

tanmayv@goa.bits-pilani.ac.in

**Abstract**

With recent technological advancements in artificial intelligence and computer vision, gait analysis is now possible on portable devices such as cell phones. However, most state-of-the-art vision-based systems still operate by enforcing many constraints for capturing a patient's video, such as the use of a static camera and maintaining a specific distance from the camera. Enforcing such constraints is possible when done under the observation of a trained professional but not in a home setting. A second problem with most vision-based systems is that the output is usually a classification label and a confidence value. The reliability of the confidence value is often questioned by medical professionals. This paper addresses these issues by presenting a novel system for gait analysis that is robust to camera movements and offers explanations for its output. The paper uses dataset containing videos of subjects wearing two types of Knee Ankle Foot Orthosis (KAFO), i.e. "Locked Knee" and "Semi-flexion" for their mobility. The dataset also contains meta-data and ground truth for the explanations. The ground truth highlights the statistical significance of seven features captured using motion capture systems in differentiating between the two gaits. To tackle the problems with camera movements, the proposed system uses super-resolution and pose estimation during its pre-processing. It then detects the seven features, i.e. Stride Length, Step Length and Duration of single support of orthotic and non-orthotic leg, Cadence and Speed, using the skeletal output of the pose estimation. The features train a multi-layer perceptron, whose output is explained, i.e., the contribution to the classification by the features is highlighted. While most state-of-the-art systems cannot even process the video or train themselves using the proposed dataset, the proposed system offers an average accuracy of 94%. The explainability of the model is also validated using the ground truth and can be considered reliable.

## 1   Introduction

The human gait comprises of bipedal movement of their body, requiring complex coordination between nervous, musculoskeletal and cardiorespiratory systems [KDCMR00]. Any pathology affecting these systems, such as neurological or systemic disorders, diseases, genetics, injuries, or ageing, can manifest as pathological gait. Thus, gait analysis has become an integral part of medical diagnosis. Traditionally, gait acquisition for medical diagnosis is performed using the "Gold standard" motion capture and force plates systems [MDLHGZMZ14]. Being expensive, cumbersome to operate, requiring a highly constrained acquisition setup and highly skilled manpower, these systems can be used only in research laboratories and hospitals. Alternatively, gait can be captured using wearable sensor-based [POS18] [PRB+22] systems or electromyographs [WSP+14]. However, these systems also require trained professionals who can attach these systems to specific parts of the body [ZWQ+17]. The recent advancements in artificial intelligence and computer vision have led to the development of several Computer Assisted Diagnosis (CAD) systems. The CAD systems for gait operate on videos captured using any cell phones or surveillance cameras and process them using machine learning models [NHFPVS+16b]. Such systems make gait-based medical diagnoses cheaper and accessible to people in remote locations. While these systems have reported high accuracy in classifying pathological gaits, medical professionals usually consider them less reliable, as they operate as a black box. Thus, this paper proposes a novel gait classification system that offers an explanation for its output. A second advantage of the proposed system is that, it can operate without enforcing any constraints on the acquisition setup. Thus, this is a significant improvement over the state-of-the-art systems that operate on videos captured using static cameras, usually mounted on a tripod stand. They also require people to maintain a specific distance from the camera and they fail to process videos captured using a non-stationary camera. The details of such state-of-the-art systems are discussed in the following sections.

## 1.1 Literature review

The lack of data is the biggest hurdle for most video-based gait analysis systems for medical diagnosis. The state-of-the-art consists of several research work that are evaluated on private datasets. These datasets are maintained private to protect the participating patients' identities and to address other ethical and privacy concerns. To the best of our knowledge, there are only five publicly available datasets for classifying pathologies. Among them, the first three contain only silhouettes of individuals, while the other two offer positions of 25 key locations (key points) on the human body.

### 1.1.1 Dataset

The first dataset, DAI [NHFPVS+16b], contains the gait of five healthy individuals represented by their silhouettes. The five individuals registered their gait three times. They then simulate different pathologies to register another fifteen sequences. The thirty videos are captured from a sagittal view, with a Kinect camera mounted three meters away from the walking individuals. The team behind the DAI dataset also published the DAI 2 dataset [NHGC17], which also contains gait from five healthy individuals. Apart from their gait, the individuals record simulation of four different pathologies, which include Parkinson's, diplegia, hemiplegia and neuropathy. It results in a total of 75 sets of silhouettes in sagittal plane, captured with a Kinect camera mounted eight meters away. The INIT dataset [OHEM18] also contains binary silhouettes belonging to nine males and one female. Every subject is recorded two times simulating eight different types of gait, which include healthy, right arm motionless, half motion in the right arm, left arm motionless, half motion of the left arm, restricted movement in the entire body, half motion of the right leg and half motion of the left leg. Thus, totalling eighty sets of silhouette sequences.
The GAIT-IT dataset [AMV+21] is captured in a studio with two synchronised cameras placed three and 1.5 meters away from the observed individuals. The first camera captures the sagittal view while the other captures the front/rear respectively. The dataset captures the same pathologies as DAI 2 but simulates two different severities for each (except healthy gait). The most significant advantage of this dataset is that it captures nineteen males and two females, totalling 828 gait sequences. The second advantage of the dataset is that it provides silhouettes and the output from OpenPose [Mar19], which contains the location of 25 key points of the individual's body. OpenPose[Mar19] is also used to create the spinal deformity dataset [SV22]. The gait is captured from the sagittal view with the camera placed three meters away from the observed individuals. The pathologies simulated in the dataset include two different severities of trunk (Kyphosis and Lordosis), along with their gait. A total of twenty-nine people participated, resulting in a total of 150 sets of key-point sequences.

### 1.1.2 Classification systems

The system that uses the above-discussed or privately captured datasets can also be classified into silhouette-based and pose-based. Most silhouette-based systems perform background subtraction [AK21] on input video during pre-processing. The resulting sequences are then used to create gait representations, such as gait energy images (GEI)[NHGC17]. These representations capture the gait motion in a single image by aggregating the silhouettes. The GEIs can be classified across different pathologies using classifiers such as support vector machine (SVM). The GEIs can also be used to define new features, such as the amount of movement and movement broadness [OHEM18], which highlight differences between different types of gait. The silhouettes can also be used to detect features such as step length, joint angles and duration of a gait cycle. These systems approximate the location of feet by selecting the bottommost right and left pixels of the silhouette [NHFPVS+16a]. Better feature values can be estimated using human anatomy information to annotate specific body parts [NHFPVS+16b]. These features are also classified using SVM. Overlapping silhouettes over a gait cycle can highlight the position of the feet when they are in complete contact with the ground. These positions can estimate step length, cadence, speed and duration of double limb support for a gait cycle, which can then be used to classify spatio-temporal parameters of gait using SVM [VSC18]. The precision of such features has been validated against a motion capture system in [VDVC+19].

The output from a motion capture system, a 3D skeletal model, is considered the gold standard in gait analysis [RV12]. It can be used to accurately estimate the joint motion of a subject undergoing knee arthroplasty to assess their recovery [WAM+17]. It can also be used to estimate features such as step length and trunk angle with sufficient accuracy to track the progression of neurological disorders, such as Parkinson's disease [JAI+15]. With the advancements in deep learning, a 2D skeletal model comprising of key points of a subject can now be estimated from an image using pose estimation models, such as OpenPose [Mar19]. These key points can be used to create gait representations, such as skeleton energy images (SEI). The SEI follows the same procedure as GEIs, aggregating the skeletons to create a single image representing the individual's gait [LC20]. The SEIs can then be used to train a shallow convolutional neural network (CNN) to classify gait [AMV+21]. Deep CNNs such as VGG119[SZ14] can also be fine-tuned using GEI or SEIs to classify gait [VCS18]. This requires each individual skeleton belonging to a gait cycle to be processed by recurrent neural networks (RNN), such as long short-term memory (LSTM) networks for better results as compared to aggregating the 2D skeleton [SV22].

## 1.2 Motivation and Contribution

The discussion in section 1.1.1 highlights the most significant disadvantage of the publically available datasets. All of them contain simulations of pathologies. The simulations are often critiqued for lack of nuances when simulating a pathological gait. The above datasets also define an acquisition setup to capture the videos. Adhering to the acquisition setups is not always feasible, especially when models trained on such datasets are used in a home setting where untrained individuals capture the video. The discussion in section 1.1.2 highlights the systems' emphasis on obtaining higher accuracies by using complex deep-learning models. The accuracy comes at the cost of explainability of the model. Thus, most medical professionals cannot draw useful conclusions using them. Training these models also requires a large amount of data. However, capturing data from pathological subjects can be challenging, which leads to the use of simulations. While the traditional systems can operate on small datasets, the features aren't accurate as they depend on the shape of the silhouettes.

This paper addresses the major problems with the state-of-the-art. It presents a dataset capturing gait of patients walking with the locked and semi-flexion knee-ankle-foot orthosis (KAFO), [Bap18]. The dataset was captured using a handheld camera in a manner that is most natural to untrained person. The dataset also includes metadata and ground truth for explainable AI (XAI) systems using a motion capture system. Furthermore, this paper proposes a novel video gait analysis system capable of performing classifications on videos from the camera which moves along with the subject. The proposed system offers an explanation for its output by highlighting the biomechanical features that contribute positively to its classification process. Another advantage of the proposed system is that it is composed of a multi-layer perception (MLP) that can be trained on small datasets. The rest of this paper is organised into three sections. Section 2 will discuss the proposed system. Section 3 will present the dataset and the evaluation of the proposed system using the proposed dataset. Section 4 will offer a conclusion and discuss possible future directions.

## 2 Methodology

The proposed system is presented in Fig. 1. It can be classified into four main modules: pre-processing, detection, classification, and explanation. The input to the system is a video sequence. The proposed system operates independently of the type or quality of the camera used. The pre-processing step uses a pose estimating neural network [BGR+20] to capture key points of the body in a normalised 2D coordinate system. To improve the accuracy of the coordinates, the proposed system enhances the quality of each frame through super-resolution [WYW+18]. The key points are then passed to the detection module that detects important bio-mechanical features, such as stride length, step length, duration of single-limb support, cadence and speed. A multi-layer perception (MLP) classifies these features according to the experimental setup. The MLP outputs the class and the probability of the input belonging to that class. To better understand the classification decision, the MLP is followed by an explanation module that highlights the features that are significant and insignificant to the results. This is advantageous to medical professionals who can use the explanation for their diagnosis rather than blindly trusting the system's classification results.

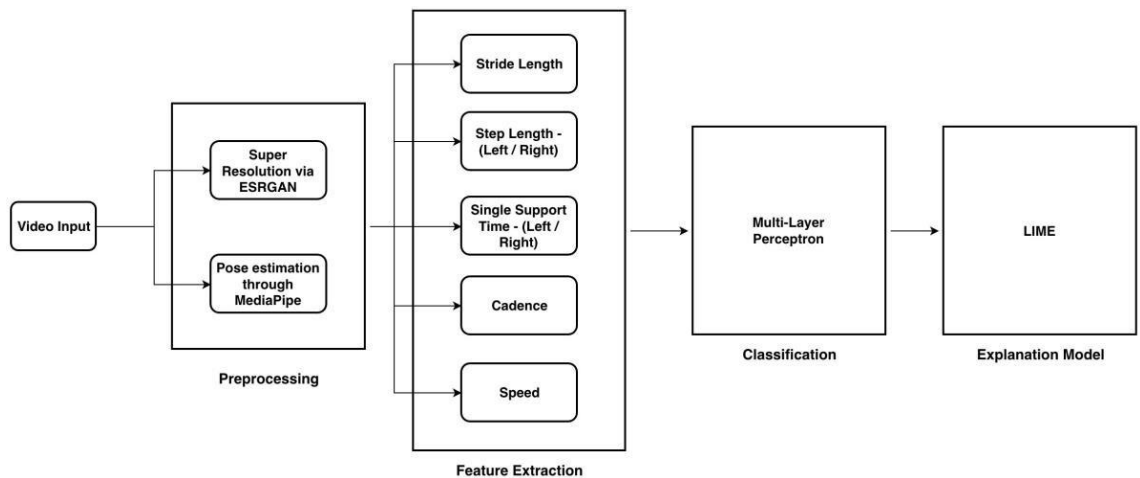

Figure 1: Architecture of the proposed system

## 2.1 Pre-processing

Given an input video, the proposed system uses the pre-processing module to convert the input video into a sequence of normalised 2D coordinates representing key points on the human body. It considers an acquisition setup without enforcing any constraints. The video quality of scenes captured in such scenarios is usually poor. Thus, direct application of the pose estimation model [BGR+20] cannot offer reliable or accurate key points. This, coupled with the fact that the KAFO occludes the leg, makes estimating the key points even more challenging. Thus, the proposed system enhances the quality of the images through super-resolution [WYW+18].

### 2.1.1 Super-resolution

The proposed system generates the super-resolution image using an Enhanced Super-resolution Generative Adversarial Network (ESRGAN) [WYW+18]. The GAN model is composed of two stages: feature extraction and image synthesis. The backbone of the feature extraction stage is a CNN called Resnet [HZRS16]. It is modified by removing batch normalisation layers and introducing a Residual-in Residual Dense Block. The deeper and more complex structure of the dense block allows the network to capture better features by connecting each layer to all the preceding layers. The discriminator of the GAN handles the image synthesis stage. The discriminator, in this case, is tweaked so that it tries to predict the probability that a real image is relatively more realistic than a fake, which allows the discriminator to capture more detailed textures. Thus, when used on the videos captured in the dataset presented in this paper, ESRGAN generates sharper video frames with better clarity and overall visual quality - see Fig. 2. The contextual understanding of ESRGAN is important in this case, as the videos in the dataset contain excessive camera movement and motion blur.

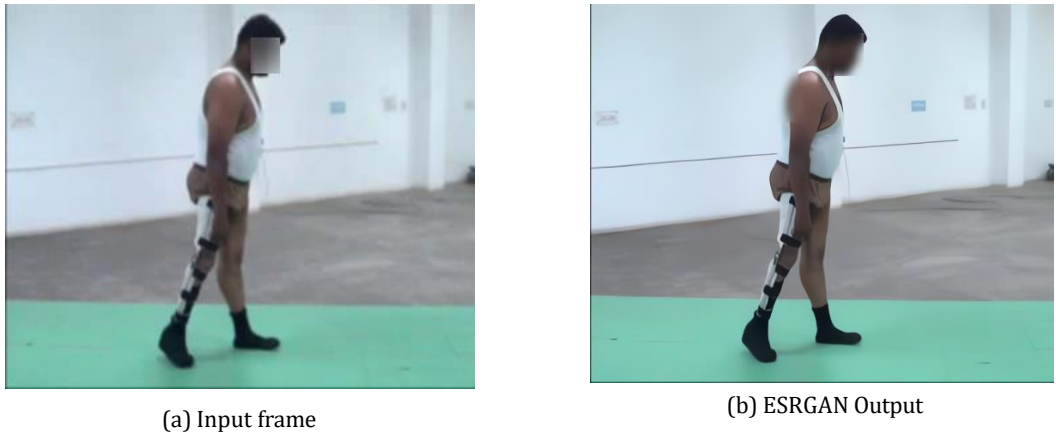

(a) Input frame  (b) ESRGAN Output

Figure 2: Comparison between frames before and after the use of ESRGAN [WYW+18]

### 2.1.2 Key point Estimation

The super-resolution frames are then used to estimate the pose of the person captured within it. Pose estimation is performed through BlazePose [BGR+20], which uses a CNN called MobileNet V2 [SHZ+18] as its backbone. It detects thirty-three key points, such as bilateral hip, knee, ankle, feet, shoulder, and other joints. These key points are defined in a topology of body parts. The network, thus, identifies a body part by estimating a small number of visible key points. The rest of the key points can be estimated through regression on the body parts. The CNN used for the detection of the key points is structured as a UNet, whose decoder estimates heatmaps for the key points given an input image. It then uses an additional regression encoder to estimate the key points' 2D coordinate locations. While the architecture of the network contributes to its results, the proposed system uses BlazePose because of its training set. Unlike other pose estimation models, such as OpenPose[Mar19], BlazePose's training set contains images of patients wearing assistive devices, such as prosthetics. This allows the network to learn to estimate the key points on assistive devices. OpenPose [Mar19] performs poorly under such scenarios as they treat locked knee as occlusion and try to estimate the key points around it -see Fig.3.

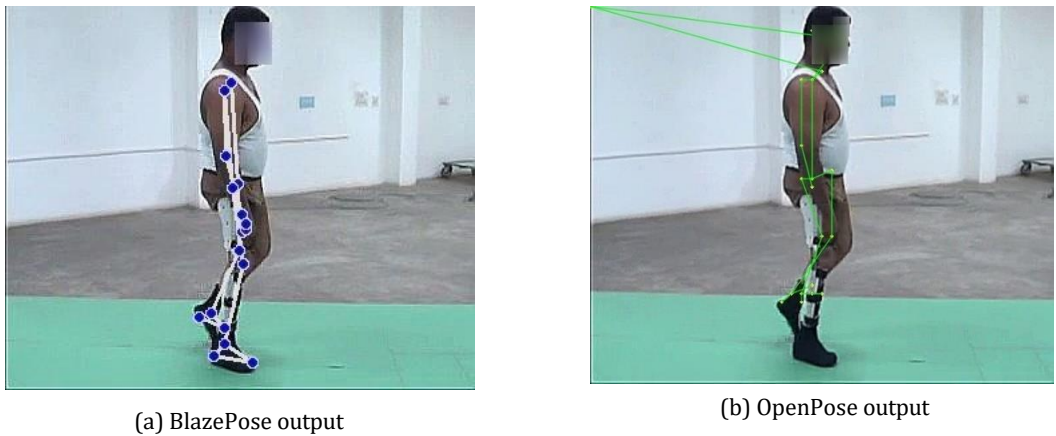

(a) BlazePose output  (b) OpenPose output

Figure 3: Comparison between BlazePose and OpenPose

## 2.2 Bio-mechanical feature detection

The output from BlazePose appears to be a skeleton composed of thirty-three key points walking in place in a (0,1) 2D coordinate space. Among its thirty-three key points, the proposed system focuses on tracking only the movement of heel key points over time. Specifically, the system uses the X-coordinate of the heels and tracks the change in its position from one frame to the next. In the (0,1) 2D coordinate space, the skeleton appears to be walking in place with the relative motion occurring between both the legs with respect to a fixed torso. The maximum separation between the limbs marks the terminal stance /pre-swing phase of one limb and terminal swing/initial contact of contralateral limb while minimum distance between the limbs marks the midstance phase of one limb and midswing phase of the opposite limb respectively. This relative movement between the limbs can be visualised as a sinusoidal wave representing the distance between the two feet, as illustrated in Fig. 4. The proposed system computes the Euclidean distance between the two heels' X-coordinates over the entire video length to capture the relative movement.

A gait cycle begins with a heel strike followed by a toe-off of the same limb. A second heel strike of the same foot marks the end of a gait cycle. In most pathological gait, especially when individuals walk with the support of orthotics, such as KAFO, patients never register a clear heel strike or a toe-off. They mostly touch the ground flat-footed. Thus, the proposed system approximates the heel strike and the toe-off of the other foot at the same point on the plot in Fig. 4. The specific points correspond to the local maxima on the plot. A gait cycle can then be estimated as frames between three local maxima on the sinusoidal wave, representing the distance between feet.

The local maxima can be easily detected in most cases where the camera is stationary. In scenarios where the camera is non-stationary, changes in perspective distort the sine wave nature of the plot. Thus, the proposed system applies a Gaussian filter to remove noise and smoothen the plot. It also uses the average distance between local maximas as a reference to remove the outliers. Once the plot is rectified, the proposed system estimates spatial, temporal and spatio-temporal features from the plot.

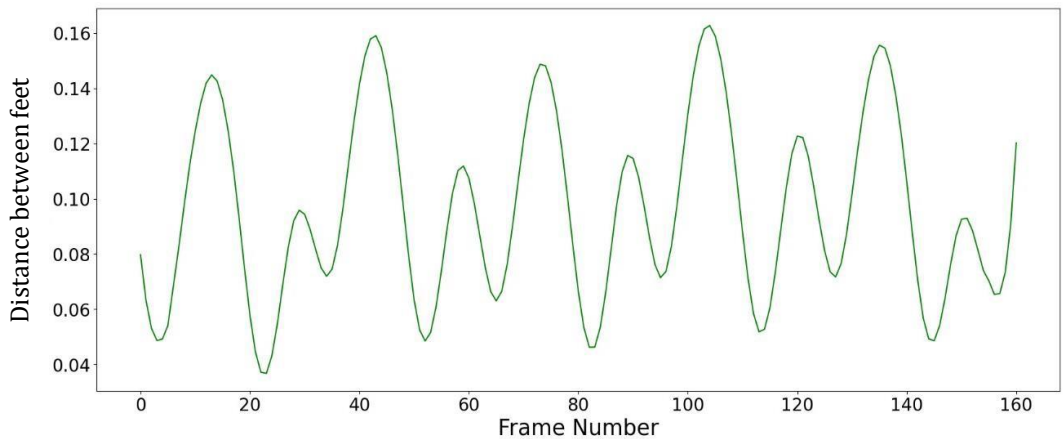

Figure 4: The plot represents the change in Euclidean distance between the feet's x-coordinates with respect to time (30 frames =1 second)

### 2.2.1 Spatial features

Step length is a fundamental parameter in gait analysis. It represents the linear distance between the heel strike of one foot and the subsequent heel strike of the opposite foot during a gait cycle. The measurement of step length corresponds to the maximum distance between the legs occurring during the double limb support phase of the gait cycle. Thus, the proposed system uses the local maxima's Y-coordinate value as the step length. Since BlazePose can track the left and the right leg, the proposed system can estimate the left and right step lengths, respectively. It should be noted that patients tend to weight-bear more on non-orthotic limbs due to instability on the other side, resulting in short step lengths. Thus, the step lengths can also be identified as the orthotic leg (OL) step length and non-orthotic leg (NOL) step length. Also, since BlazePose normalises the key points, the step length in meters can be calculated as follows:

$$Step\ length_{meters} = Height_{meters} \times Step\ length_{normalized} \qquad (1)$$

The next spatial feature the proposed system considers is the stride length. It is defined as the distance between successive points of heel strike of the same foot. The proposed system estimates the stride length by adding two consecutive step lengths within a gait cycle. Thus, similar to step length, the proposed system computes the left and right stride lengths for a gait cycle. However, unlike the step length, the two stride lengths are usually equal. Thus, the proposed system outputs an average stride length for a gait cycle.

### 2.2.2 Temporal features

The only temporal features the proposed system considers is the duration of single limb support. It represents the duration of the gait cycle for which one foot is in contact with the ground and the other is in the swing phase. Similar to step length, the duration of single limb support can be calculated for both the left and right leg. However, the proposed system considers annotating the duration of single limb support bilaterally and annotations follow from step length.

The proposed system estimates the duration of single limb support using the x-axis plot value illustrated in Fig. 4. The local maxima on the plot represents the heel strike and toe-off of the other foot. Thus, the difference between the two consecutive local maxima's x-coordinate values can be used to estimate the duration of single limb support. Since the resulting value represents the time in frames, the proposed system converts the time in seconds using the frame rate of the camera, following equation 2:

$$Duration\ of\ single\ support_{sec} = \frac{1}{frame\ rate} \times Duration\ of\ single\ support_{no.\ of\ frames} \qquad (2)$$

### 2.2.3 Spatio-temporal features

The final set of features considered by the system are speed and cadence. The speed of a person refers to the rate at which the individual is walking, typically measured in meters per second (m/s). The proposed system calculates the speed for each gait cycle. The distance an individual covers is represented by the sum of the maxima of three consecutive Y-coordinate values, as illustrated in Fig. 4. The difference in the X-coordinates of the maximas can estimate the time. The resulting value can be converted into meters per second using the height and frame rate of the camera, similar to equations 1 and 2. Thus, the speed of an individual is computed as follows:

$$speed_{m/s} = Height_{meters} \times framerate \times \frac{\sum_{i=1}^{3} maxima_i}{frame_3 - frame_1} \qquad (3)$$

The final feature considered by the proposed system is cadence. It represents the rate at which individuals complete their strides, typically measured in steps per minute. Similar to speed, the proposed system estimates the steps using the plot in Fig. 4. Each local maxima in the figure represents a step, with three steps composing a gait cycle. The cadence can be estimated by extrapolating the number of steps an individual will take for one minute. It is computed by the proposed system as follows:

$$cadence_{steps/minute} = \frac{3 \times 60 \times framerate}{(frame_3 - frame_1)} \qquad (4)$$

This paper's set of features is limited to seven because of the dataset considered for evaluations in section 4. The dataset contains videos and corresponding ground truth features captured using a motion capture system [Bap18].

## 2.3 Multi-layer perceptron (MLP)

The proposed system performs classification using an MLP. Considering the complexity of the classification task and the amount of data available, the MLP is designed to have an input layer, two hidden layers, and an output layer. The input layer contains a perceptron for each input feature. Each hidden layer is composed of 64 perceptrons with a Rectified Linear Unit (ReLU) activation function. The final output layer contains perceptrons equal to the number of classes with a SoftMax activation function. The training of the network is performed using an Adam optimiser and cross-entropy loss function. The model is set to a learning rate of $10^{-5}$ with the number of epochs set to 1000, with early stopping. The output of the MLP is the class label and confidence in its classification. It does not communicate any information about which features influence the classification process, similar to classifiers such as KNN and SVM used by state-of-the-art systems. To obtain an explanation for the classification, the proposed system uses an explanation module called Local Interpretable Model-agnostic Explanations (LIME) [RSG16]. It offers insights into feature importance, contributing to the transparency which is essential in assessing complex gait analysis in medicine.

## 2.4 Explainability

The proposed system uses LIME[RSG16] to offer an explanation for its classification. The explanation highlights the features that contribute most to the MLP's output. It is achieved by creating simpler, interpretable models that approximate the result of the MLP. LIME begins by creating an instance of the dataset, which consists of an individual or a set of features. Once an instance is created, LIME augments it to create a new dataset. It then uses a kernel function that measures similarity to assign weight to the new dataset based on its similarity with the initial instance. Finally, it trains a surrogate model, typically a linear regression or decision tree, on the weighted dataset. The purpose of the surrogate model is to approximate the complex MLP model's behaviour while being interpretable. Once trained, the surrogate model can offer explanations for the complex MLP model by analysing its coefficients and identifying the features that contribute most to the final classification. These features can be validated with the ground truth available in the dataset to verify the system's reliability.

# 3 Evaluation

The proposed system is evaluated on a new dataset presented in this section. The protocol considered for the evaluation is stratified k-fold cross-validation. It is performed by dividing the dataset into five folds, where the first four folds are used for training, and the last fold is used for validation. This process is repeated until all five folds are used to validate the model. The evaluation metric of average accuracy across the five folds is used to report the results. To avoid user bias, the training and validation sets are mutually exclusive with respect to the participants.

## 3.1 The KAFO user Dataset

The paper presents original dataset collected from the users of KAFO for the evaluation of the proposed model. It should be noted that most publicly available datasets, as discussed in section 1, are simulated. They are often critiqued for the quality of their simulation, as healthy individuals cannot imitate the nuances of a pathological gait. The proposed dataset consists of six patients walking with the support of two different KAFO models, "Locked Knee" and "Semi-flexion". The use of Semi-flexion KAFO allows stance phase knee flexion particularly in the initial contact and loading response phase of the gait cycle. Thus, the energy spent while walking is considerably reduced, leading to a more comfortable walk as compared to the locked knee orthosis [Bap18]. Unlike other acquisition setups for the datasets discussed in Section 1.1.1, no instructions were offered to the trained medical professional recording the subject's gait. The gait analyst captured the gait in a way that was most natural. The resulting videos are thus captured using a handheld camera from a mostly sagittal viewpoint. The videos contain a lot of motion blur as the gait analyst walks along with the subject. The distance between them often shifts as the gait analyst recording the video sometimes moves towards and at other times, away from the subject. The gait analyst also sometimes overtakes the patient while at other times lags behind the patient. The above acquisition setup best depicts real-world scenarios where no constraints can be enforced on capturing a video, as illustrated in Fig. 5. The six patients register three gait cycles per video to create a dataset with two classes, "Locked Knee KAFO" or KAFO1 and "Semi-flexion KAFO" or KAFO2. The dataset also contains additional information, such as the patients' meta-data, which includes height and ground truth for the seven features discussed in section 2. The ground truth is constructed using a PhaseSpaceTM (PhaseSpace Inc, San Leandro, CA, USA) motion capture system comprising active LED markers and eight infrared cameras [Bap18] and was conducted at Movement Analysis Laboratory, Department of Physical Medicine and Rehabilitation, Christian Medical College, Vellore. The original study was approved by the institutional review board and ethical committee of the Christian Medical College and Hospital (CMC), Vellore, India. All subjects signed written informed consent before participating in the study.

The first experiment involved training the MLP to classify the gait as KAFO-1 or KAFO-2. It follows the above-discussed stratified k-fold cross-validation for the evaluations. To increase the size of the dataset, each gait cycle is clipped out and considered a separate data point. Due to the camera's movement, environmental factors and the nature of gait, each gait cycle contains sufficient variation to act as a unique data point. The augmentation increases the dataset's size three-fold. The increased size of the dataset prevents the MLP from overfitting. It should be noted that some gait cycles contain excessive motion, distorting all useful information. Thus, gait cycles that do not return an output during pre-processing are discarded. The output from the pre-processing is used to detect the seven features: Stride Length, Step Length (both OL and NOL), duration of single limb support (both OL and NOL), Cadence and Speed.

## 3.2 Results

The classification is performed over these features using the proposed system, and the results are reported in Table 1.

Table 1: Results of k-fold cross-validation classifying KAFO1 and KAFO2

| System | Accuracy |
| --- | --- |
| Proposed Features+SVM [VSC18] | 83% |
| Proposed system (without super-resolution) | 87% |
| Proposed system | 94% |

Apart from reporting the results of the proposed system, Table 1 also compares the results with the state-of-the-art system [VSC18]. It should be noted that the state-of-the-art system [VSC18] fails at detecting features, as their pre-processing cannot handle the challenging dataset. Problems are also encountered when deep learning models-based systems such as [VCS18] and [SV22] are trained with the dataset. The low amount of training data leads to overfitting. Thus, to compare with the state-of-the-art, the features from the proposed system are used as input to the system presented in [VSC18]. The classification is then performed using SVM.

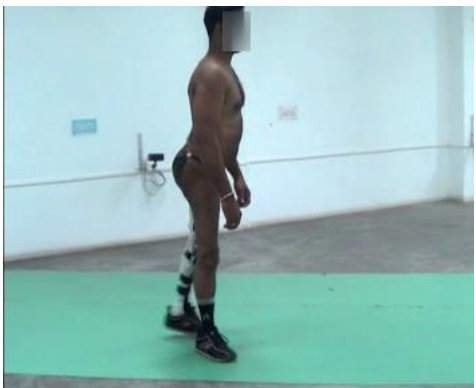
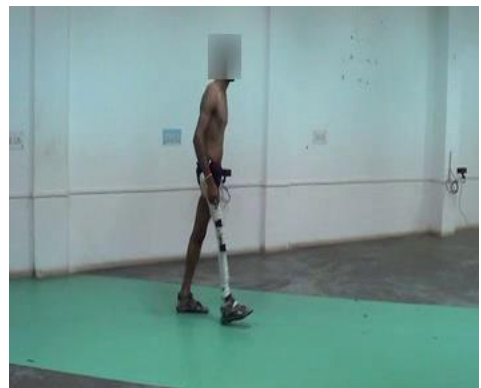

(a)             (b)

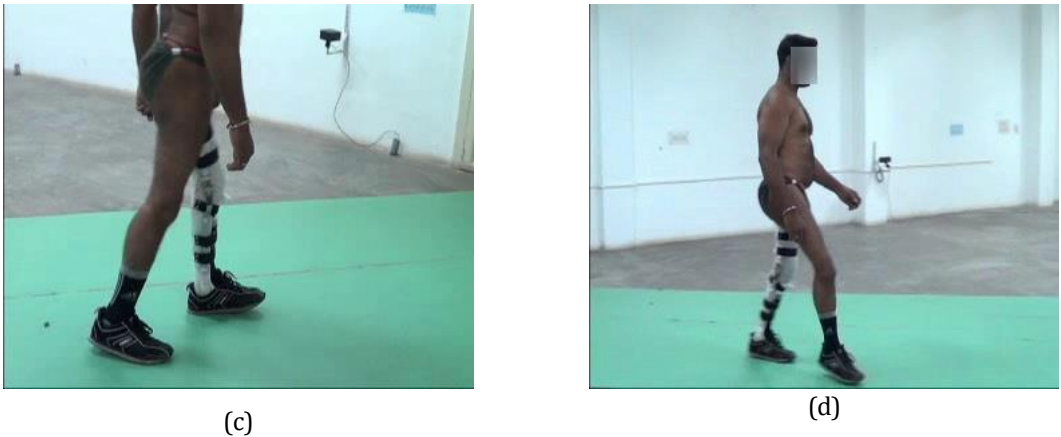

(c)           (d)

Figure 5: Frames captured at different camera positions while recording videos for the dataset.

Table 2: *p*-values for features belonging to KAFO1 and KAFO2 captured using the motion capture system

| Features | Step length (OL) | Step length (NOL) | Average Stride length | Duration of single support (OL) | Duration of single support (NOL) | Cadence | Speed |
|---|---|---|---|---|---|---|---|
| *p*-Value | 0.758 | 0.866 | 0.674 | 0.901 | 0.291 | 0.112 | 0.186 |

The results suggest that the proposed system is better at classifying the two gaits than the state-of-the-art system [VSC18]. It can also be concluded from the results that each part of the pre-processing step is integral in obtaining the results, as skipping super-resolution or using alternate pose estimation models, such as OpenPose [Mar19], can significantly affect the results of the system. The proposed system comes with the added benefit of explaining its output. In Fig. 6 the positive count represents the number of times the feature leads to correct classification, and the negative count represents the number of times the feature did not contribute to the classification. Thus, for correct classifications, the features that contribute most to correct decisions are the cadence and duration of single limb support (NOL). They exhibit a high count of positive contributions for correct classifications. Hence, they can be considered significant in obtaining the classification results. It can be attributed to the fact that the patient's weight was more on NOL than the OL, which was successfully captured by the proposed system. Other features, such as step length (NOL), stride length, and speed, display a balanced distribution of positive and negative count, i.e. partially contributing to the classification results. Finally, features such as the step length (OL) and duration of single limb support (OL) yield more counts of negative contributions to the classification results and can be considered insignificant. Clinically, similar justifications could be thought about less weight-bearing OL.

The explanations can also be validated using the previously published biomechanical study performed in [BapG18]. All the features from section 2 are captured using a combination of motion capture in [BapG18] and act as the ground truth for the proposed system. Table 2 reports the *p*-values, which indicate the probability of observing differences between KAFO1 and KAFO2 features due to chance. The *p*-values support the explanation with cadence, duration of single support (NOL), and speed being among the features with low p-values. It indicates that the differences between KAFO1 and KAFO2 are less likely due to chance with respect to those features. Meanwhile, the differences between the step length (NOL), step length (OL), and duration of single support (OL) are statistically insignificant across KAFO1 and KAFO2.

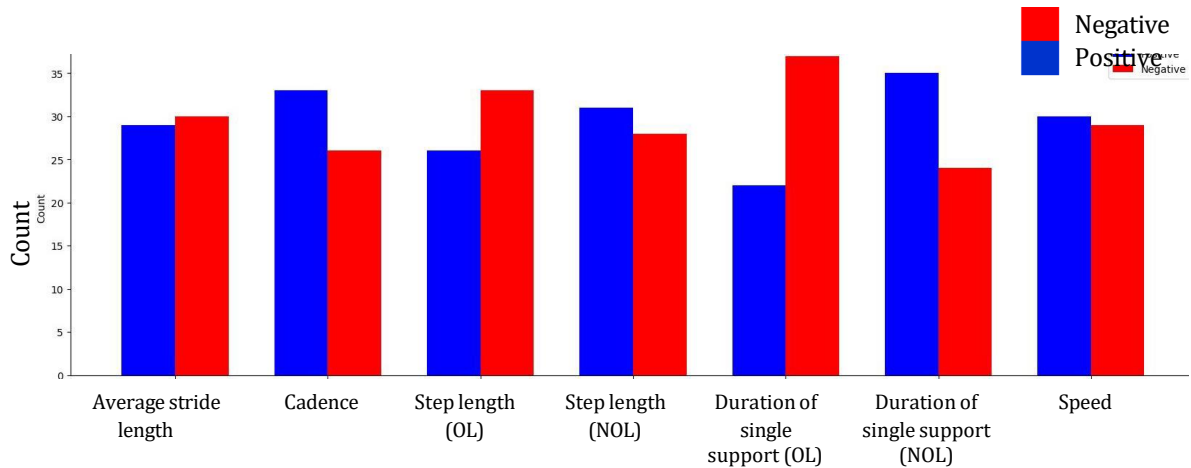

Figure 6: Representing the instances when a feature contributes positively and instances when the features do not contribute (indicated as negative) to the correct classification

Fig. 6 illustrates the contribution of individual features in classifying the validation set. However, for a given input, multiple features contribute towards its classification. The MLP outputs the label and the confidence value. The explanation module then outputs the explanation as reported in Table 3.

Table 3: Comparison between correctly classified and misclassified explanation

| Features | Misclassified as KAFO 1 with a confidence of 62% | Correctly classified as KAFO 1 with a confidence of 99% |
|---|---|---|
| Average Stride Length | 0.06 | 0.07 |
| Cadence | 0 | 0.20 |
| Step Length (OL) | 0.05 | 0.08 |
| Step Length (NOL) | 0.01 | 0.11 |
| Duration of Single Support (OL) | 0.02 | 0 |
| Duration of Single Support (NOL) | 0.04 | 0.03 |
| Speed | 0 | 0.30 |

Table 3 presents two examples: the first is misclassified as KAFO 1, and the second is classified correctly as KAFO 1. In either case, the output of the MLP is KAFO 1 andonly the confidence differs. Traditional ML models do not offer any other details beyond these values. The proposed system offers an importance score for each feature. In Table 3, column 1, important features for the classification are average stride length and step length (OL), while cadence does not support the classification. From Table 2 and Fig. 6, it can be inferred that the stride length and step length (OL) are seldom evidence for correct classification. In contrast, cadence, a good feature for classification between KAFO 1 and KAFO 2, does not support the classification. Thus, the output of the MLP can be considered unreliable. In contrast, Table 3. Column 2, the classification is supported by reliable features such as cadence, step length (NOL), and speed. Thus, the output label from the proposed system can be trusted.

## 4 Conclusion

The paper presents an XAI-based gait analysis system for patients walking with Knee-Ankle-Foot orthosis. The system is designed to work in unconstrained settings, such as when videos are captured using handheld cameras, which is the most natural way to capture videos for naï ve individuals. To the best of our knowledge, no state-of-the-art exists that considers this scenario. They operate with well-defined acquisition setups, where the camera is mounted on a static tripod stand. The movement of the camera causes challenges, such as excessive motion blur, cropping of body parts and changes in view of the people. The paper also focuses on patients wearing orthosis, which occludes the legs. In all these cases, the state-of-the-art systems fail. The proposed system addresses these problems by using super-resolution to improve the video quality. It then uses BlazePose to estimate normalised key points. BlazePose operates significantly better than other pose estimation models, such as OpenPose, on patients wearing orthotics because of its training set. It then detects biomechanical features using the normalised key points, which are classified using an MLP with an accuracy of 94 %. The proposed system also offers explanations for its classification using LIME.

This paper uses dataset of patients walking with the support of two different KAFO models, Locked Knee and Semi-flexion. The dataset also contains ground truth for the explanation, reported as the p-value for the following features: Step length (OL), Step length (NOL), Stride length, duration of single support (OL), duration of single support (NOL), Cadence and Speed, captured using a motion capture system. The results suggest that the explanation of the models aligns with ground truth, highlighting the reliability of the proposed system.

While augmentation presented in this paper allows the training of the MLP, a larger dataset is needed to use deep learning models. Thus, capturing a larger dataset can be a possible future direction. The deeper models can then replace the proposed system's detector module. With neural network-based detectors, the system will learn important features, significantly improving its performance.

## Acknowledgements

We thank the original study participants and the Institutional Review Board and Ethics Committee of CMC Vellore for approving the study. We are grateful to Mrs. Joyce for her help in the gait and video data collection. We also thank Dr. Prashant Chalagiri, Assoc. Prof., Dept. of PMR, CMC Vellore, who provided clinical support in the study and Dr. S. Sujatha, Dept. of Mech Eng, IIT Madras, Chennai, India, who guided the initial work. The work is supported by BITS Pilani, Goa Campus, research initiation grant.

## Further Material

The dataset used for the classification has been made publicly available at : http://tinyurl.com/5ds5f33c